%% file: main.tex
\documentclass[10pt,twocolumn,letterpaper]{article}

\usepackage{iccv}              
\usepackage{microtype}
\usepackage{graphicx}
\usepackage{caption}
\usepackage{booktabs} 
\usepackage[textsize=tiny]{todonotes}
\usepackage{amsmath}
\usepackage{amssymb}
\usepackage{mathtools}
\usepackage{amsthm}
\usepackage{makecell}

\input{preamble}

\definecolor{iccvblue}{rgb}{0.21,0.49,0.74}
\usepackage[pagebackref,breaklinks,colorlinks,allcolors=iccvblue]{hyperref}

\def\paperID{13511} 
\def\confName{ICCV}
\def\confYear{2025}

\title{Lifelong Learning with Task-Specific Adaptation: Addressing the Stability-Plasticity Dilemma}

\author{
Ruiyu Wang$^1$, Sen Wang$^2$, Xinxin Zuo$^3$, Qiang Sun$^1$ \\
$^1$University of Toronto, $^2$HoYoverse, $^3$Concordia University \\
{\tt\small rwang@cs.toronto.edu, qiang.sun@utoronto.ca}
}

\begin{document}
\maketitle
\begin{abstract}  
Lifelong learning (LL) aims to continuously acquire new knowledge while retaining  previously learned knowledge. A central challenge in LL is the stability-plasticity dilemma, which requires models to balance the preservation of previous knowledge (stability) with the ability to learn new tasks (plasticity). While parameter-efficient fine-tuning (PEFT) has been widely adopted in large language models, its application to lifelong learning remains underexplored. To bridge this gap, this paper proposes AdaLL, an adapter-based framework designed to  address the dilemma through a simple, universal, and effective strategy. AdaLL co-trains the backbone network and adapters under  regularization constraints, enabling the backbone to capture task-invariant features while allowing the adapters to specialize in  task-specific information. 
Unlike methods that freeze the backbone network, AdaLL incrementally enhances the backbone's  capabilities across tasks while minimizing interference through  backbone regularization. This architectural design significantly improves both stability and plasticity, effectively eliminating the stability-plasticity dilemma. Extensive  experiments demonstrate that AdaLL consistently outperforms existing methods across various configurations, including dataset choices, task sequences, and task scales.
\end{abstract}
\section{Introduction}

In real-world applications, data often arrive sequentially, in batches, or through periodic updates. These dynamics are further complicated by factors such as limited storage capacity and privacy concerns. In such scenarios, traditional concurrent or multitask learning approaches, which depend on training models on a single, static, and large dataset, become impractical. Lifelong learning (LL), also known as Incremental Learning (IL) or Continual Learning (CL) \footnote{We use these terms interchangeably in this paper.}, is specifically designed to address these challenges in dynamic environments. This paradigm enables models to continuously acquire new knowledge from incoming data without relying on previously seen data. Unlike traditional methods that require access to the entire dataset for training, incremental learning allows models to adapt over time to new tasks using only the data of the current task while preserving knowledge from previous tasks.

An ideal lifelong learning algorithm must effectively address the trade-offs between stability and plasticity. Specifically, it should balance the retention of previously learned knowledge with the acquisition of new knowledge. This challenge, referred to as the stability-plasticity dilemma, is a fundamental problem in incremental learning \citep{mermillod2013stability}. Particularly, directly optimizing the model for the new tasks often results in a significant loss of accuracy, which is known as catastrophic forgetting. Additionally, efficient memory utilization is crucial, requiring either no storage or  minimal retention of data points from previous tasks. This requirement becomes particularly critical in  real-world scenarios, where  transient data and constraints related to privacy and memory resources are prevalent.

Elastic Weight Consolidation (EWC) \citep{kirkpatrick2017overcoming} and Learning without Forgetting (LwF) \citep{li2017LwF} were among the first methods to address the stability-plasticity dilemma, focusing on mitigating catastrophic forgetting. In recent years, Low-Rank Adaptation (LoRA) \citep{hu2021lora, liang2024inflorainterferencefreelowrankadaptation, zhang2020sidetuningbaselinenetworkadaptation} have demonstrated promising performance in mitigating catastrophic forgetting.

However, vanilla LoRA \citep{hu2021lora} struggles to maintain stability in incremental learning due to significant interference across  tasks. To mitigate this issue, InfLoRA \citep{liang2024inflorainterferencefreelowrankadaptation} initializes and constrains LoRA adapter parameters for new tasks within a subspace orthogonal to the gradient subspace of previously learned tasks.
Despite its advantages, InfLoRA has three major limitations. First, the orthogonal subspace diminishes over tasks, limiting plasticity improvement. Second, InfLoRA relies on prior knowledge of task numbers to predefine the number of LoRA modules, reducing its scalability or applicability in complex  scenarios where the number of tasks is large or unknown. Third, InfLoRA requires a pretrained model, restricting its use in settings where such a model is unavailable.
Therefore, a new lifelong learning algorithm is needed -- one that effectively utilizes adapters while addressing the stability-plasticity dilemma.



\textbf{Contributions} 
To address these challenges, this paper proposes a simple, effective, and universal \textbf{Ada}pter-based \textbf{L}ifelong \textbf{L}earning framework, referred to as \textbf{AdaLL}.  AdaLL integrates task-specific adapters at the top of the backbone and co-trains the backbone network and adapters under regularization constraints, enabling the backbone to capture task-invariant features while the adapters specialize in task-specific information. Unlike methods that freeze the backbone network, AdaLL incrementally improves the backbone's ability across tasks by leveraging the growing sample size, while minimizing interference through backbone regularization. This novel approach improves both stability and plasticity simultaneously, effectively eliminating the stability-plasticity dilemma. Additionally, our framework eliminates the need to store gradient subspace derived from previous tasks, making it more memory efficient than InfLoRA.  Last, our framework is able to integrate seamlessly with other incremental learning algorithms such as EWC \citep{kirkpatrick2017overcoming}, LwF \citep{li2017LwF},  and  DualPrompt \citep{wang2022dualpromptcomplementarypromptingrehearsalfree}. 
Extensive experiments demonstrate that AdaLL, when integrated with state-of-the-art algorithms, consistently outperforms their counterparts across various configurations, including dataset choices, task sequences, and task scales.

\section{Related work}
\begin{figure*}[t!]
    \centering
    \includegraphics[width=0.95\textwidth]{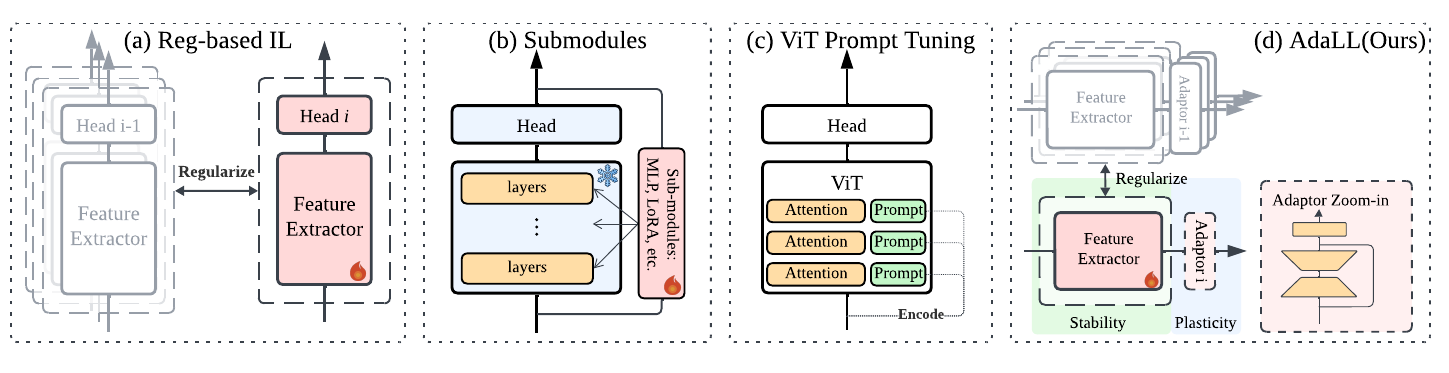}
    \caption{Existing methods contribute to incremental learning (IL) in various ways: (a) Regularization-based methods, such as LwF and EWC,  introduce regularization constraints to preserve knowledge from previous tasks; (b) Submodule-based approaches, for instance InfLoRA and SideTuning, integrate additional components such as MLPs and LoRA into the backbone network to improve adaptability while freezing the backbone to maintain stability; and (c) Prompt tuning methods (DualPrompt, CodaPrompt, etc.) introduces task-specific prefixes to the key and value in attention modules to improve task-specific performance. Our framework (d) uses submodules in a way that it can be benefited from regularization and other backbone-specific algorithms to ensure a better response to the stability-plasticity dilemma.} 
    \label{fig:methods-overview}
\end{figure*}

This section describes related works. Two primary strategies are commonly employed to address catastrophic forgetting in incremental learning problems: regularization-based and submodule-based methods.

\textbf{Regularization-based IL.} Regularization-based methods mitigate catastrophic forgetting via introducing additional regularization constraints. They attempt to preserve the prior knowledge via explicit regularization, weight-selection, or rehearsal\footnote{Parameter-selection and rehearsal can be treated as some implicit regularization methods.}. Explicit regularization-based methods mitigate forgetting by regularizing differences in weights or output predictions between the old and new models \citep{kirkpatrick2017overcoming, zenke2017Continual, aljundi2018memory, li2017LwF, dhar2019learning, joseph2022energy}. Weight-selection-based methods mitigate forgetting  by identifying and freezing weights that are critical to previously learned tasks while optimizing the other weights \citep{mallya2018packnet, serra2018overcoming, wang2022learningpromptcontinuallearning, rajasegaran2020itamlincrementaltaskagnosticmetalearning}. Rehearsal-based methods preserve prior knowledge by retaining instances from previous tasks and training on a combined dataset that includes these instances alongside data from the current task. These retained data may consist of exemplar images \citep{rebuffi2017icarl, chaudhry2018riemannian}, publicly available external datasets \citep{lee2019overcoming, zhang2020class}, or synthetic data generated by generative models for the old tasks \citep{shin2017continual, kemker2017fearnet, he2018exemplar}.


\textbf{Submodule-based IL.}
Submodule-based  methods introduce additional submodules or subnetworks to mitigate catastrophic forgetting  \citep{zhang2020sidetuningbaselinenetworkadaptation, liang2024inflorainterferencefreelowrankadaptation, bhat2023taskawareinformationroutingcommon, liang2024inflorainterferencefreelowrankadaptation, pham2021dualnetcontinuallearningfast}. These methods freeze the backbone and fine-tune only the added subnetworks.  The architectural designs of these submodules vary. For example, TAMiL \citep{bhat2023taskawareinformationroutingcommon} employs an attention-based subnetwork, whereas some other methods utilize LoRAs. Prompt-tuning methods \citep{wang2022learningpromptcontinuallearning, smith2023coda, wang2022dualprompt, wang2022s} can also been seen as submodule-based  methods that learn  a prompt pool to properly instruct a pre-trained  model to learn different tasks. However, prompt-tuning methods require image tokens being processed through the network twice, leading to increased inference latency. Moreover, current prompt-tuning methods are tailored to ViT or attention modules, making them incompatible with other robust architectures such as ResNet.

In contrast to all previous works, our framework combines the strengths of both regularization-based methods and submodule-based approaches, where regularizing the backbone improves stability while task-specific adapters enhance plasticity for adapting to new tasks. Additionally, our framework is universal, allowing seamless integration with various architectures, and does not require prior task knowledge which hinders scalability.



\section{Incremental learning with task-specific adaptation}

\subsection{What is missing in existing approaches?}

As shown in Figure \ref{fig:methods-overview}, Most existing methods focus on improving either stability or plasticity, addressing only one side of the stability-plasticity dilemma. For example, regularization-based methods aim to enhance stability by introducing additional constraints, while submodule-based methods \citep{zhang2020sidetuningbaselinenetworkadaptation} improve plasticity by incorporating submodules, often at the cost of freezing the entire backbone network to preserve stability. However, this approach not only compromises plasticity but also weakens stability, as a backbone pretrained on other tasks does not necessarily capture task-invariant information in the new scenarios, limiting its ability to generalize across tasks. In summary, stability and plasticity appear to be in conflict, giving rise to the so-called stability-plasticity dilemma.

To further investigate this dilemma, we consider a classification problem in incremental learning. A typical classification model compromises a backbone network and a classifier or prediction head for each task. We argue that this dilemma arises from the conventional perspective of treating the model as a single, monolithic block. Optimizing the entire model for new tasks leads to catastrophic forgetting, whereas freezing the backbone limits plasticity. In response, we propose a two-block network design for incremental learning, where stability and plasticity are explicitly assigned to separate components. Specifically, we conceptualize the model as a two-block structure: the backbone network functions as an invariant feature extractor shared across all tasks, while each classifier head serves as a task-specific module. This architecture is shown in Figure \ref{fig:methods-overview} (d).

The two-block design is aimed to resolve catastrophic forgetting. This problem occurs when the backbone is optimized solely for the current task, while freezing the backbone results in insufficient plasticity for learning new tasks. To introduce sufficient plasticity without compromising stability, adapters are integrated as feature modifiers on top of the backbone. While the backbone's primary effort lies in learning invariant features, adapters provide the necessary flexibility to accommodate new tasks. Furthermore, to ensure the backbone captures invariant features, we impose backbone regularization which encourages the similarity between previously learned and newly acquired features. Unlike freezing the backbone, regularizing it allows the model to leverage the increasing sample size across tasks, enabling the incremental learning of invariant representations.

Our framework is universal. In contrast to prompt-based methods \citep{wang2022dualprompt, wang2022learningpromptcontinuallearning, smith2023codapromptcontinualdecomposedattentionbased}, which work exclusively with ViTs, it is not backbone-specific. Moreover, classical regularization-based methods such as EWC \citep{kirkpatrick2017overcoming} and LwF \citep{li2017LwF}, as well as more recent approaches, can be seamlessly integrated into our framework to enhance performance when used for backbone regularization. Additionally, our method does not require task information before training, offering greater scalability compared to InfLoRA \citep{liang2024inflorainterferencefreelowrankadaptation}. We discuss the design of adapters in Section \ref{sec:adapter-def} and the modification of regularizations in Section \ref{sec:adapter_integration}.

\subsection{Introducing adapters} \label{sec:adapter-def}

\begin{figure*}[h!]
    \centering
    \includegraphics[width=0.8\textwidth]{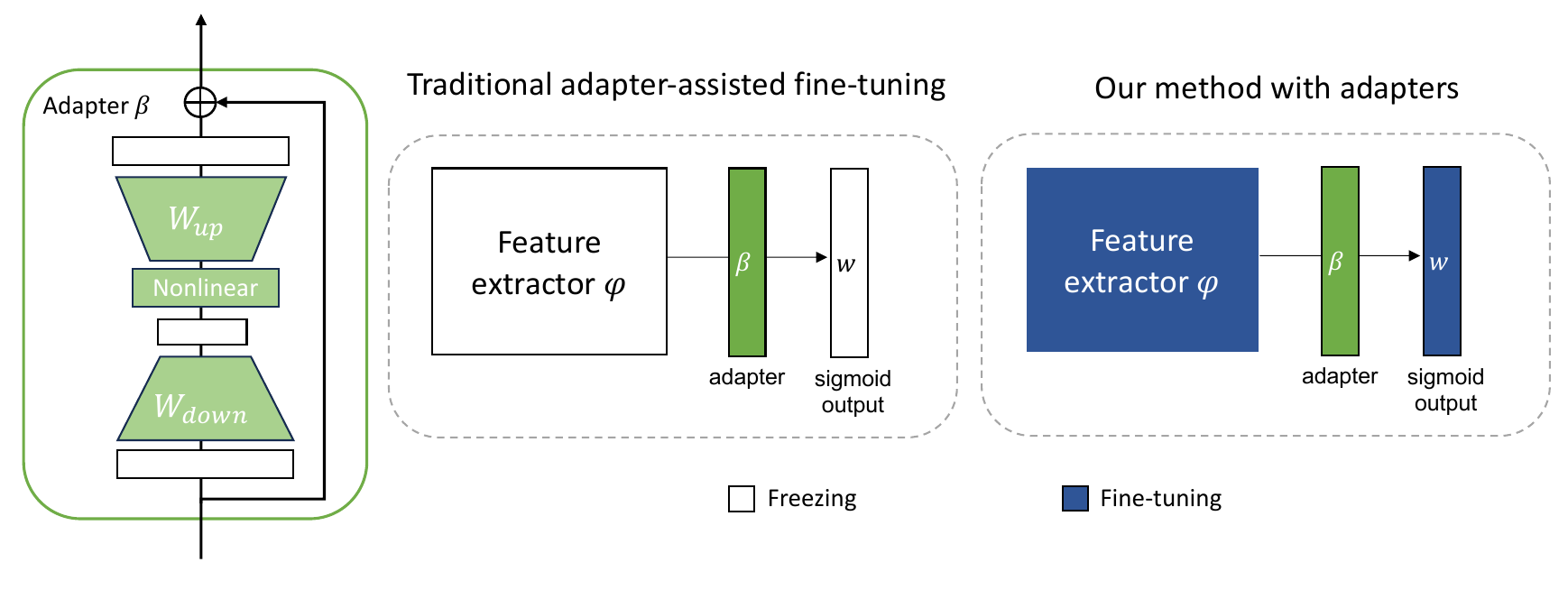}
    \caption{Architecture of the adapter and a comparison highlighting the distinctions in its implementation between traditional fine-tuning and our method. Left: an adapter consists of the down-projection, the nonlinear transformation, up-projection, and skip-connection. Right: The key difference between traditional use of adapter and ours is that we co-train adapter with the entire network when learning a new task.}\label{fig:fig_graphical_illustration}
\end{figure*}

The adapters are positioned between the backbone network, denoted as $\varphi$,  and the prediction layer, serving as task-specific feature modifiers $\beta^t$ for each task $t$. As illustrated in Figure \ref{fig:fig_graphical_illustration}, the adapters follow a conventional bottleneck structure. Starting with an initial dimension $d$ and a bottleneck width $b$, the down-projection layer reduces dimensionality from $d$ to $b$ using a fully-connected layer with a weight matrix $\mathbf{W}_{d \times b}$ and a non-linear activation function $g$, expressed as 
\$
\texttt{Down}_{d\rightarrow b}(\mathbf{x}) = g\left( \mathbf{x} \mathbf{W}_{d \times b} \right).
\$
Similarly, the up-projection layer is defined as
\$
\texttt{Up}_{b\rightarrow d}(\mathbf{x}) = g\left( \mathbf{x} \mathbf{W}_{b \times d} \right).
\$

Each adapter consists of both a down-projection and an up-projection layer, connected to the output via a skip-connection \citep{he2021towards}. This bottleneck design allows the adapter to utilize both backbone features from $\varphi(x)$ and the modified features processed through the down- and up-projection layers:
$$
\beta^t(\varphi(x)) = \varphi(x) + \texttt{Up}_{b\rightarrow d}^t \left( \texttt{Down}_{d\rightarrow b}^t \left( \varphi(x)\right) \right). 
$$
For each task, a new adapter is initialized randomly. The inputs are processed sequentially by the feature extractor, the task-specific adapter, and the classifier head, which transforms the feature representation into class predictions. The training process and loss computation are determined by the specific training methods integrated into our framework.

Our adapter module incorporates three key advantages: 1. It harnesses the strengths of PEFT modules, which have demonstrated exceptional performance in enabling language models to efficiently learn new features. 2. It mirrors classical model architectures when treating the inserted adapter as part of the classifier head. This design allows both regularization-based and prompt-tuning methods to be seamlessly integrated into our framework. 3. By enabling the backbone and adapters to capture task-invariant and task-specific information separately, our approach effectively addresses the stability-plasticity dilemma. This can be further enhanced through additional regularization techniques, one of which we introduce in the next section to facilitate invariant learning.

\subsection{Learning invariant features by backbone regularization}\label{sec:adapter_integration}

As discussed previously, we aim to regularize the backbone to facilitate the learning of task-invariant features while allowing the adapters to specialize in task-specific information. To illustrate this approach, we start by modifying prediction-regularized and weight-regularized methods, integrating adapters and backbone regularization. 

\vspace{-8pt}
\paragraph{Prediction regularization}


Prediction-regularization methods, such as LwF and Learning without Memorizing (LwM) \citep{dhar2019learning}, mitigate forgetting by aligning the predictions of the old and new models through regularization. In addition to the task loss, model outputs at task $t$ are regularized using a distillation loss that aligns them with the outputs from all previous tasks $t'$, where $1\leq t'< t$. The overall loss at task $t$ is:
\begin{align*}
    \mathcal{L}^t &= \ell^t(\theta) + \lambda_{\text{distill}} R^t_{\text{distill}} \\
    &=\ell^t(\theta) + \lambda_{\text{distill}} \sum_{t'=1}^{t-1}  M \left(\varphi^{t'} (x), \varphi^{t} (x) \right),
\end{align*}
where $M$ is a similarity metric such as cosine similarity or cross-entropy, and $\lambda_{\text{distill}}$ is a hyperparameter that controls the strength of the distillation loss.

However, while the backbone network is expected to learn task-invariant features and the adapters to capture task-specific information, the above formulation does not explicitly enforce this separation. To encourage invariant feature learning in the backbone, we introduce the following backbone regularization term:
\$
R^t_{\varphi} = \sum_{t'=1}^{t-1} M\left( \text{Linear}_{d\times c}(\varphi^{t'}(x)), \text{Linear}_{d\times c}(\varphi^{t}(x)) \right),
\$
where $c\leq d$ is a reduced dimension. In practice, we set $c$ to the number of classes for each task, as this choice intuitively makes the regularization act as a direct distillation on the backbone network. 

To learn a new task $t$, we define a new loss function that combines the task loss, the distillation term, and the backbone regularizer to align the backbone across tasks:
\$
    \mathcal{L}_t = \ell_t(\theta) + \lambda_{\text{distill}} R^t_{\text{distill}} + \lambda_{\varphi} R^t_{\varphi},
\$
where $\lambda_{\varphi}$ is a hyperparameter that controls the strength of the backbone regularization,  and $\lambda_{\text{distill}}$ balances the retention of prior knowledge with adaptation to new tasks in the incremental learning process.  We integrate this backbone regularizer into LwF, as it has demonstrated strong performance in incremetnal learning tasks based on our experience. 

\paragraph{Weight regularization}

Weight-regularization-based methods, such as EWC \citep{kirkpatrick2017overcoming}, directly regularize the neural network's weights. At each task $t$, EWC combines the task loss with a weight regularization term: 
\$
\mathcal{L}^t = \ell^t(\theta) + \sum_{t'=1}^{t-1}\sum_{i} \frac{\lambda}{2}F_i(\theta_i - \theta^*_{t', i})^2,
\$
where $\theta$ represents the current model's weights, $F_i$ is the $i$-th diagonal entry of the Fisher information matrix $F$, and $\theta^*_{t', i}$ is the $i$-th parameter for task $t'$. 

The regularization term provides stability. To improve plasticity without compromising stability, we propose regularizing only the backbone's weights while leaving the adapters unconstrained. Let $ A$ denote the index set for adapter weights. The new loss for task $t$ is defined as 
\$
\mathcal{L}^t = \ell^t(\theta) + \sum_{t'=1}^{t-1}\sum_{i\not\in A} \frac{\lambda}{2}F_i(\theta_i - \theta^*_{t', i})^2.
\$
This modification can be applied to all weight-regularized methods. For example, both Memory Aware Synapses \citep{aljundi2018memory} and Path Integral \citep{zenke2017Continual} can incorporate the adjustments.

\section{Experiments}\label{sec:num_studies}
This section quantitatively evaluates our framework. Specifically, we examine its impact on incremental learning, focusing on \textbf{efficacy} and \textbf{universality}. To this end, we address the following questions:
\begin{enumerate}
    \item Does AdaLL enhance incremental learning performance? This is analyzed in Section \ref{sec:comprehensive-comparison}.
    \item Is AdaLL universally effective across different incremental learning algorithms, and does it improve their performance? We explore this in Section \ref{sec:classic-methods}.
    \item How do different framework configurations influence incremental learning, and how does AdaLL handle challenges such as hyperparameter selection? These aspects are discussed in the ablation studies (Section \ref{sec:ablation}).
\end{enumerate}
    
\noindent We begin by outlining the experimental setup.

\subsection{Experimental setup}

\paragraph{Datasets}
We evaluate the performance of different methods on two datasets: CIFAR-100 \citep{krizhevsky2009learning} and ImageNet \citep{russakovsky2015imagenet}. CIFAR-100 consists of images with a resolution of $32 \times 32 \times 3$ and serves as our primary dataset for analyzing the impact of adapters across various settings\footnote{For ViT on CIFAR-100, we resize the image size to $224 \times 224 \times 3$.}. Additionally, we include ImageNet, which offers a more diverse set of training images with a resolution of $224 \times 224 \times 3$. To mitigate training time and computational constraints, we limit our analysis to the first 100 classes of ImageNet. A summary of dataset statistics is provided in Appendix \ref{appendix:sec:dataset}.

Task ordering is an important but often overlooked aspect of dataset (Appendix \ref{apdx:task-ordering}). We explore its impact on CIFAR-100 by exploring not only the default alphabetical but also the iCaRL and coarse ordering. The iCaRL ordering is a random ordering with a fixed seed used by iCARL \citep{rebuffi2017icarl}, and the coarse ordering groups similar classes within tasks based on the 20 coarse categories of CIFAR-100. If not specified, the alphabetical ordering is adopted.

\paragraph{Network architectures}

Following \citep{de2022continual_survey, hou2019learning}, we adopt different architectures for the two datasets. For CIFAR-100, we use ResNet-34, while for ImageNet, we follow \citep{resnet} and use ResNet-18. Unless otherwise specified, we do not use pretrained weights for ResNet models.

Recent studies highlight the advancements of transformer-based architectures in continual learning. For methods that utilize ViTs as the backbone \citep{wang2022dualprompt, liang2024inflorainterferencefreelowrankadaptation}, we follow their choice of \texttt{vit-base-patch16-224-in21k} with pretrained weights to ensure a fair comparison.

Our training and hyperparameter configurations are included in Appendix \ref{appendix:sec:implementation}.

\paragraph{Evaluation metrics}

We evaluate task-incremental learning (Task-IL) performance across our experiments. This evaluation assesses each approach’s performance when task identifiers (task-IDs) are provided. We adopt this protocol as our framework is particularly impactful in a multi-head setting, where a task-ID oracle identifies the appropriate task-specific head during inference. In contrast, class-incremental learning (Class-IL) does not provide such information, making it a more practical yet challenging scenario. Since Class-IL is not our primary focus, we include these results in Appendix \ref{appendix:sec:tag} to explore AdaLL's universality.

To compare overall learning performance across different methods, we compute the average top-1 accuracy at each task $t$, defined as:
$
A_t = \frac{1}{t} \sum_{i=1}^T a_{t, i},
$
where $a_{t, k}$ is the accuracy on task $k$ after training on task $t$. For efficacy evaluation, we report the accuracy after training the final task and the average accuracy across all tasks. For universality evaluation, we report the average performance over 10 runs with different random seeds for both CIFAR-100 and ImageNet.

\paragraph{Baselines}
We evaluate a variety of baseline methods to examine whether integrating or modifying AdaLL adapter setups improves \textbf{efficacy}. Our evaluation includes methods originally designed for ResNet, such as EWC \citep{kirkpatrick2017overcoming}, LwF \citep{li2017LwF}, and iTAML \citep{rajasegaran2020itamlincrementaltaskagnosticmetalearning}, as well as ViT-based methods such as DualPrompt \citep{wang2022dualprompt} and InfLoRA \citep{liang2024inflorainterferencefreelowrankadaptation}. Additionally, we migrate InfLoRA to ResNet, as it is the only method utilizing LoRA modules, allowing us to explore its broader impact beyond the ViT backbone.

To assess AdaLL’s \textbf{universality}, we evaluate its performance on a diverse set of backbone regularization approaches. Specifically, we compare AdaLL adapters with their non-adapter counterparts in weight-regularized methods, including EWC \citep{kirkpatrick2017overcoming}, Memory Aware Synapses (MAS) \citep{aljundi2018memory}, and Path Integral (PathInt) \citep{chaudhry2018riemannian}, as well as in prediction-regularized methods such as LwF \citep{li2017LwF} and Learning without Memorizing (LwM) \citep{dhar2019learning}.

\begin{table}[h]
    \centering
    \begin{tabular}{ l l l }
    \Xhline{2\arrayrulewidth}
    \textbf{Method Name} & \textbf{Avg. Acc.} & \textbf{Final Acc.}\\
    \hline
    \textit{ResNet backbones}\\
    \hspace{10px} EWC & \hspace{6px} 54.8 & \hspace{6px} 50.8\\
    \hspace{10px} EWC-A & \hspace{6px} {55.6}\color{red}{$_{+1.2}$} & \hspace{6px} {52.7}\color{red}{$_{+2.9}$}\\
    \hspace{10px} LwF & \hspace{6px} 70.9 & \hspace{6px} 71.5\\
    \hspace{10px} LwF-A & \hspace{6px} {73.8}\color{red}{$_{+2.9}$} & \hspace{6px} {72.3}\color{red}{$_{+0.8}$} \\
    \hspace{10px} iTAML & \hspace{6px} 79.0 & \hspace{6px} 80.5 \\
    \hspace{10px} iTAML-A & \hspace{6px} {84.1}\color{red}{$_{+5.1}$} & \hspace{6px} {80.6}\color{red}{$_{+0.1}$}\\
    
    \textit{ViT backbones} \\
    \hspace{10px} DualPrompt & \hspace{6px} 88.2 & \hspace{6px} 86.3\\
    \hspace{10px} DualPrompt-A & \hspace{6px} {89.3}\color{red}{$_{+1.1}$} & \hspace{6px} {87.9}\color{red}{$_{+1.6}$}\\
    \hspace{10px} InfLoRA & \hspace{6px} 91.6 & \hspace{6px} 86.7\\
    \hspace{10px} LwF-A & \hspace{6px} 90.9 & \hspace{6px} 86.4\\

    \Xhline{2\arrayrulewidth}
    \end{tabular}
    \captionof{table}{Comparison of algorithms on CIFAR-100. Methods tagged with \texttt{-A} indicate the integration of AdaLL adapters. \textbf{Avg. Acc.} and \textbf{Final Acc.} denotes the 10-task average accuracy and the last-task accuracy, respectively. For each pair with and without adapter, accuracy improvement exceeding 1\% is highlighted. }
    \label{tab:method-benchmark}
\end{table}

\begin{figure*}[h]
    \centering
    \includegraphics[width=\textwidth]{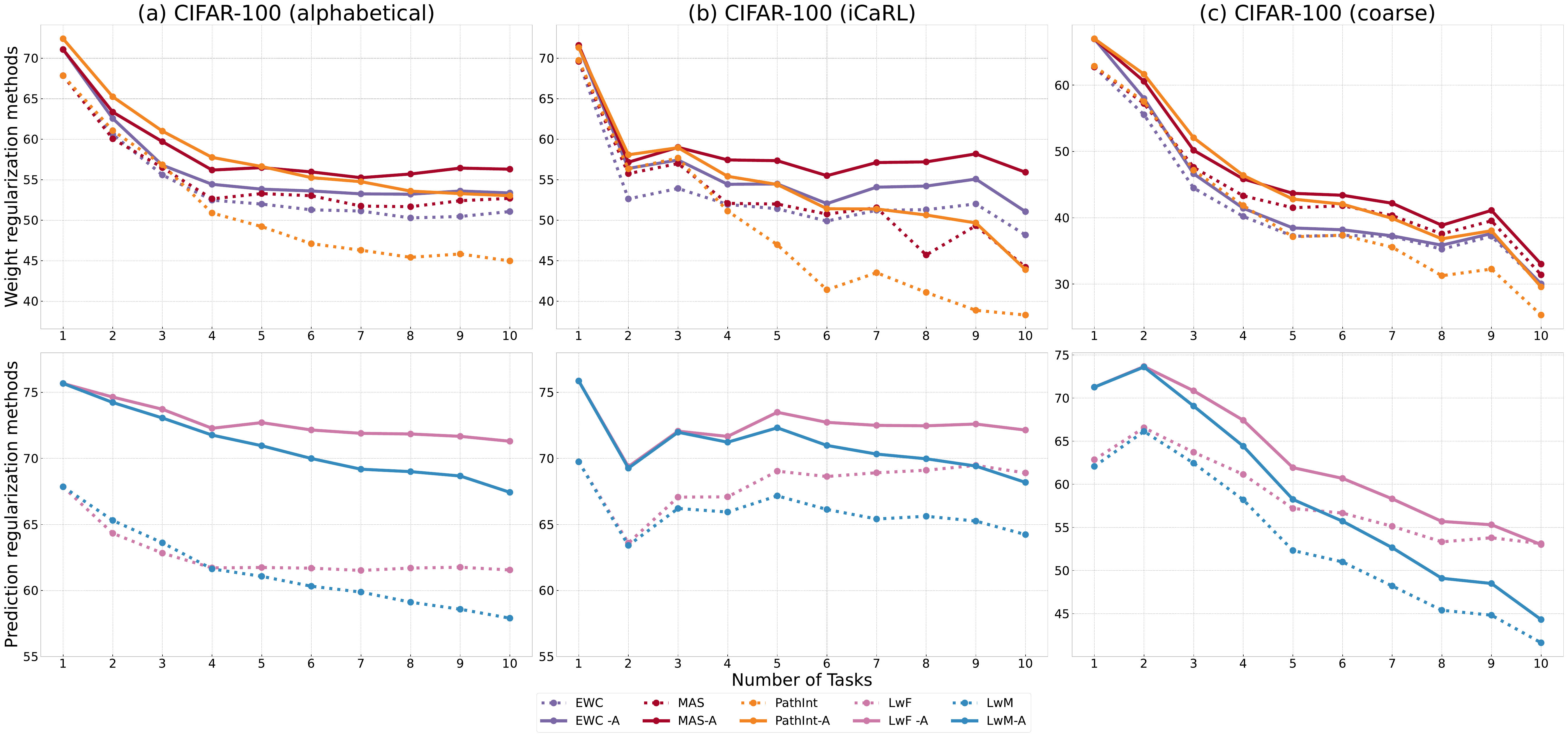}
    \label{fig:ordering_cifar}
    \caption{10-seed average accuracy for methods with or without adapters on different orderings of CIFAR-100 in Task-IL. From left to right: alphabetical order, iCaRL order and coarse order. The x-axis represent the number of tasks and the y axis represents the TOP-1 accuracy (\%). The solid line represents the results with adapter, while the dashed line represents the results without adapter.}
    \label{fig:ordering_comp}
\end{figure*}

\begin{figure*}[h]
    \centering
    \includegraphics[width=\textwidth]{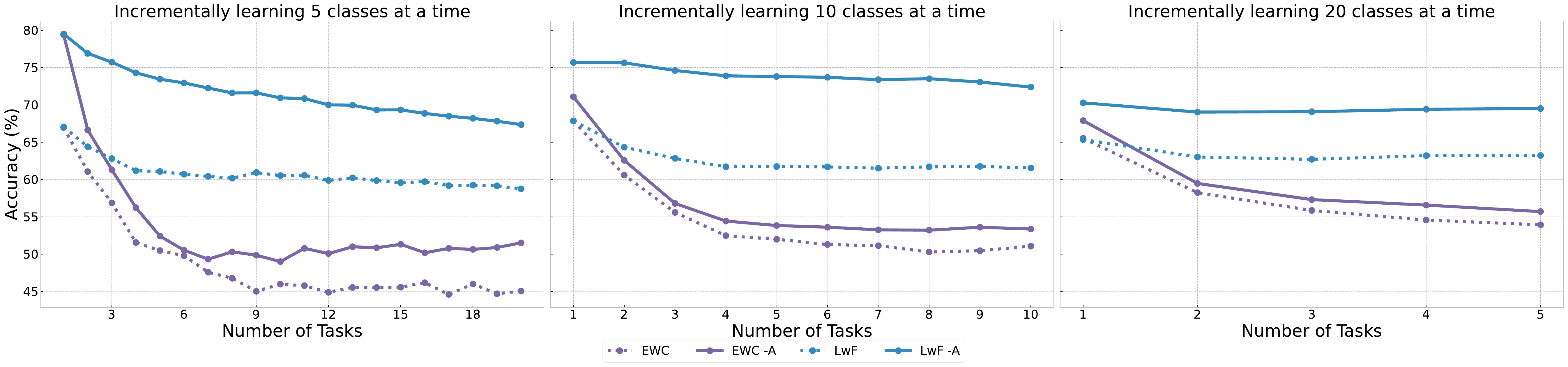}
    \caption{10-seed average accuracy for EWC and LwF with learning 5, 10, and 20 classes at a time on CIFAR-100 (alphabetical) in Task-IL. The TOP-1 accuracy is reported, with the solid lines and dashed lines represent the results with and without adapters respectively.}
    \label{fig:fig_data_scale}
\end{figure*}

\subsection{Evaluation on Efficacy} \label{sec:comprehensive-comparison}

In this section, we assess the efficacy of AdaLL by comparing it against both widely recognized and state-of-the-art incremental learning algorithms. For methods that do not incorporate a LoRA module, we integrate AdaLL to evaluate its impact on performance. For methods that already leverage LoRA, we replace their LoRA configurations with our adapter setup to examine potential improvements. 

The results are presented in Table \ref{tab:method-benchmark}. Among ResNet backbones, the AdaLL-enhanced methods (EWC-A, LwF-A, and iTAML-A) consistently outperform their vanilla counterparts with improvement around 2\% in both averaged accuracy and final-task performance. For ViT backbones, DualPrompt-A only gains a slight improvement over DualPrompt (+1.1 avg and + 1.6 final). While Lwf-A follows closely to InfLoRA with $<1\%$ difference, the impact with the two methods are further discussed using ablation studies in Section \ref{sec:ablation}. Overall, these results highlight that AdaLL improves baseline methods and consistently ranks among the top-performed methods.


\subsection{Evaluation on Universality} \label{sec:classic-methods}

\begin{figure*}[h]
    \centering
    \includegraphics[width=0.8\textwidth]{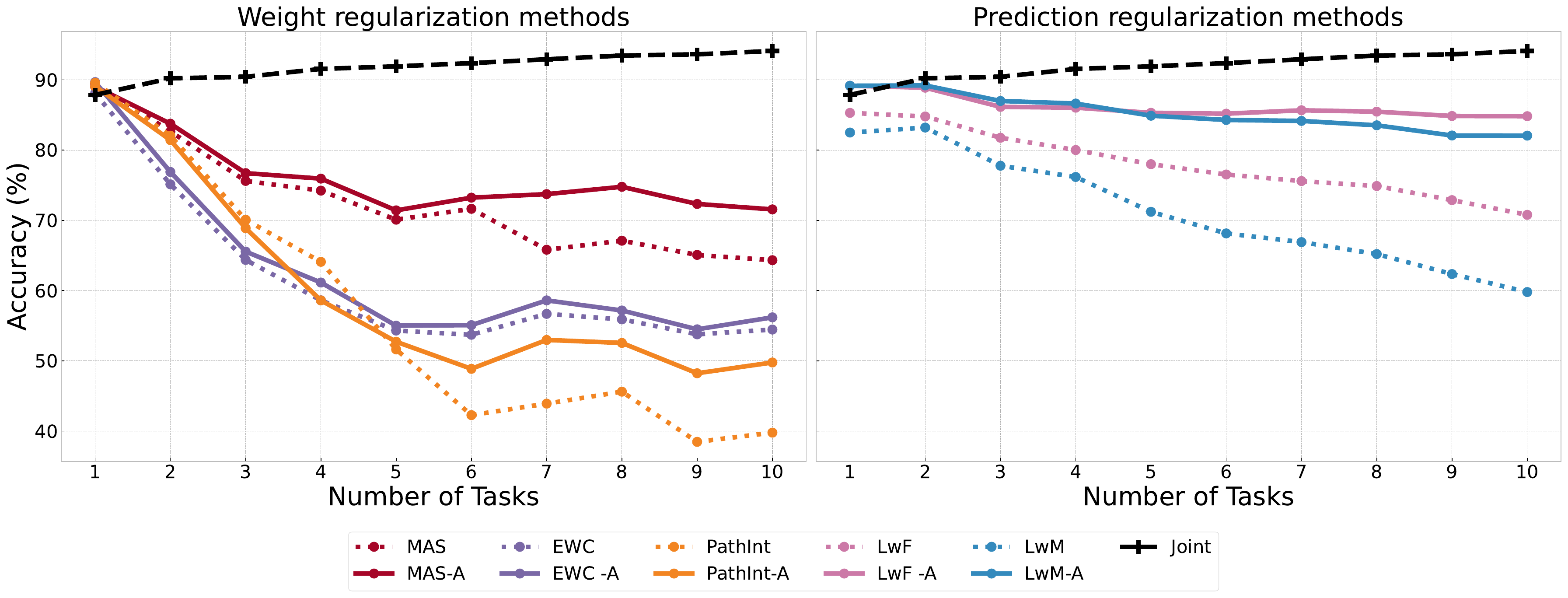}
    \caption{10-seed average accuracy for regularization methods on ImageNet-subset. 
    The x-axis represent the number of tasks and the y axis represents the TOP-1 accuracy (\%). The solid line represents the results with adapter, while the dashed line represents the results without adapter. The black, dashed line denotes the result of joint training.}
    \label{fig:imgnet}
\end{figure*}

In this section, we assess the universality of AdaLL, i.e., its consistency in delivering superior performance across different settings. Since AdaLL addresses the stability-plasticity dilemma through its two-block architecture and regularization mechanisms, we hypothesize that its benefits are not algorithm-specific. This suggests that AdaLL should enhance any method aligned with the rationale presented in Section \ref{sec:adapter_integration}.

To evaluate this hypothesis, we examine AdaLL across four key aspects: (1) a broader range of regularization method variants, (2) different class orderings, as discussed in Appendix \ref{apdx:task-ordering}, (3) varying task sizes, and (4) a larger dataset (ImageNet-subset).

The results, presented in Figures \ref{fig:ordering_comp}, \ref{fig:fig_data_scale}, and \ref{fig:imgnet}, indicate that AdaLL with weight regularization consistently outperform EWC. This advantage persists across different task orderings and sizes. On the ImageNet-subset, the performance gap narrows but remains evident, as, Figure \ref{fig:imgnet} shows that LwM-A performs closest to joint training, which serves as the upper bound for incremental learning algorithms. These findings underscore the strong potential of AdaLL in effectively addressing the challenges of continual learning across diverse settings.

\subsection{Ablation Studies} \label{sec:ablation}
\begin{figure}[t]
\centering
\includegraphics[width=0.8\linewidth]{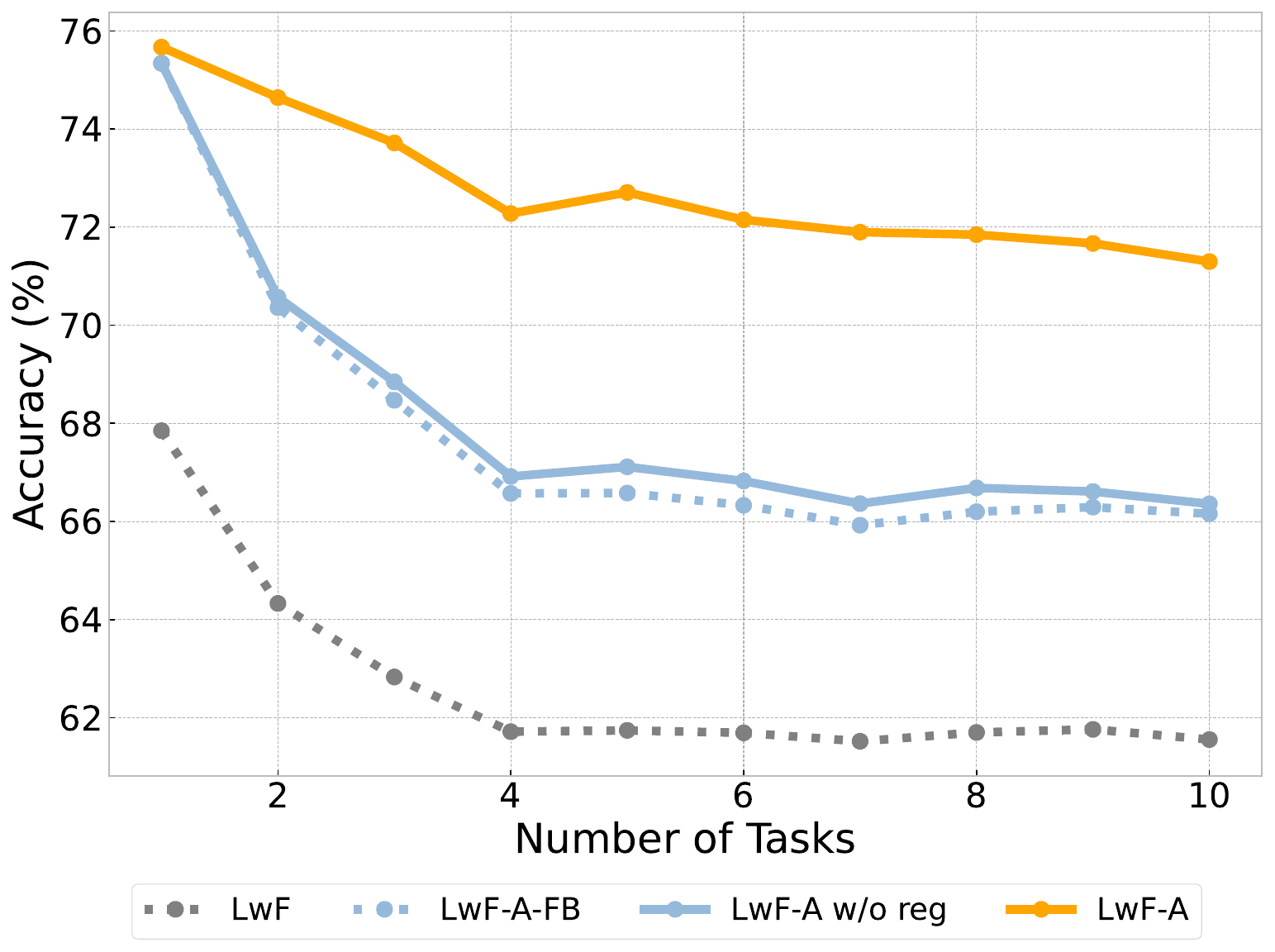}
\captionof{figure}{Impact of co-training vs. frozen backbone on CIFAR-100. We evaluate LwF with different configurations: \texttt{A} indicates using AdaLL adapters, \texttt{FB} represents a frozen backbone, and \texttt{w/o reg} removes backbone regularization as detailed in Section \ref{sec:adapter_integration}.}
\label{fig:freeze-backbone}
\end{figure}

\paragraph{Should we freeze the backbone parameters?}\label{sec:freeze-backbone}

\begin{figure*}[t!]
    \centering
    \includegraphics[width=0.8\textwidth]{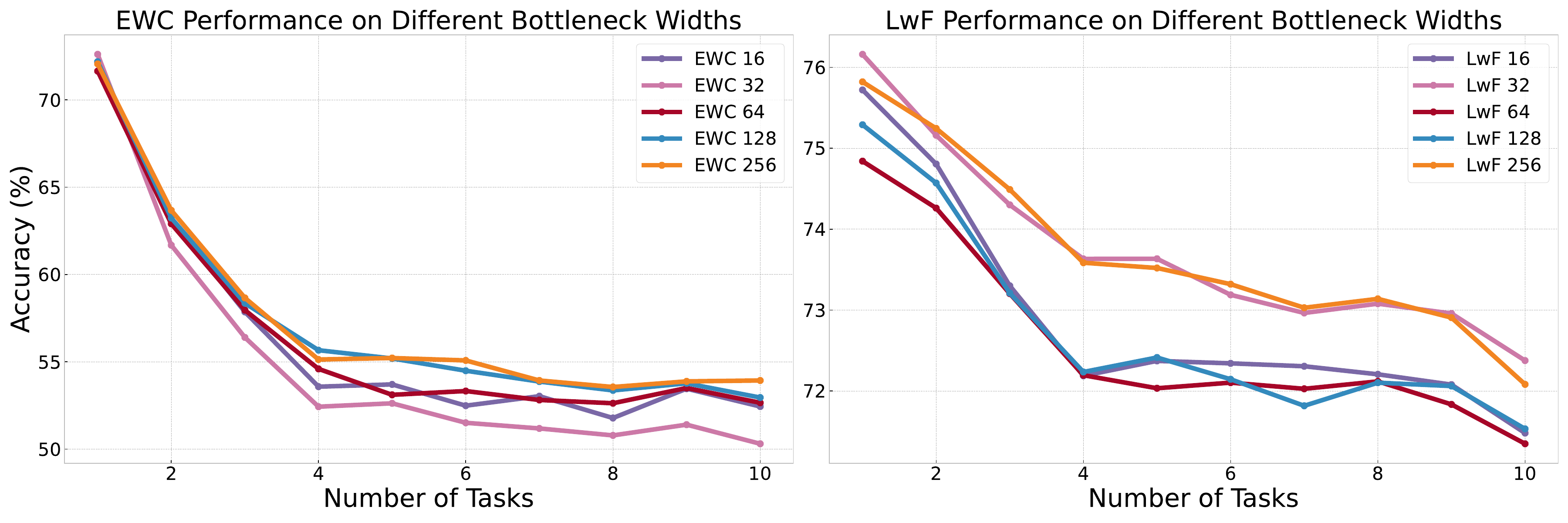}
    \label{fig:bottleneck_width_comp}
    \caption{The performance of EWC and LwF methods with different adapter bottleneck width choice on the CIFAR-100 dataset (alphabetical ordering) in Task-IL.
    The suffix \texttt{16/32/64/128/256} indicates the method implemented with width 16, 32, 64, 128, and 256, respectively.}
    \label{fig:param_comp}
\end{figure*}

Recent studies propose training submodules while keeping the backbone frozen \citep{zhang2020sidetuningbaselinenetworkadaptation, liang2024inflorainterferencefreelowrankadaptation}. However, AdaLL co-trains both the backbone and adapters, as we conjecture this, with the help of backbone regularization, enables the backbone to learn task-invariant features across tasks. To validate this, we conduct ablation studies using LwF \citep{li2017LwF} as the base algorithm and compare four configurations: 1. Baseline LwF (LwF), 2. LwF with adapters and a frozen backbone (LwF-A-FB), 3. LwF with adapters but without backbone regularization (LwF-A w/o reg), and 4. LwF with both adapters and backbone regularization (LwF-A). Figure \ref{fig:freeze-backbone} presents the results, and we have the following observations:
\begin{enumerate}
\item Adding adapters improves baseline LwF performance, while backbone regularization significantly mitigates forgetting, as seen in the slower accuracy decline over tasks.
\item The model with a co-trained backbone and regularization outperforms the frozen-backbone counterpart.
\item The model with adapters but no backbone regularization performs comparably or slightly better than the frozen-backbone model.
\end{enumerate}

These findings confirm that freezing the backbone limits the model’s ability to leverage the increasing sample size across tasks. The difference is marginal at the beginning, as no explicit constraint enforces task-invariant feature integration; but it becomes evident once more tasks are available. 

\paragraph{What about LoRA?}
\begin{table}[h]
    \centering
    \begin{tabular}{ l l l }
    \Xhline{2\arrayrulewidth}
    \textbf{Method Name} & \textbf{Avg. Acc.} & \textbf{Final Acc.}\\
    \hline
    \hspace{10px} InfLoRA-ResNet* & \hspace{6px} 42.1 & \hspace{6px} 39.4\\
    \hspace{10px} LwF-A-ResNet* & \hspace{6px} 71.2 & \hspace{6px} 58.6\\
    \hspace{10px} InfLoRA-ViT$^\dagger$ & \hspace{6px} 19.4 & \hspace{6px} 10.3\\
    \hspace{10px} LwF-A-ViT$^\dagger$ & \hspace{6px} 22.3 & \hspace{9px} 9.3\\
    \Xhline{2\arrayrulewidth}
    \end{tabular}
    \captionof{table}{Comparison of algorithms on CIFAR-100. Methods tagged with \texttt{-A} indicate the integration of AdaLL adapters. Postfix ResNet* and ViT$^\dagger$ denotes the methods using ResNet with pretrained weights and ViT with tunable parameters, respectively.}
    \label{tab:special-comp}
\end{table}

Our previous findings highlight the trade-offs between freezing and training the backbone. However, LoRA modules are commonly integrated into the backbone while keeping other network components frozen. InfLoRA \citep{liang2024inflorainterferencefreelowrankadaptation}, inspired by standard LoRA, applies this approach to ViTs, maintaining a frozen backbone and training only LoRA modules.

 InfLoRA performs slightly better than  our methods.  However, InfLoRA utilize a pretrained ViT model, whereas our methods are trained from scratch. This advantage likely stems from the pretrained model rather than InfLoRA's the architectural design, making direct comparisons unfair. Therefore, we consider two additional scenarios\footnote{Appendix \ref{appendix:sec:implementation-detail} details all InfLoRA and LwF setups for fair comparison.}:
\begin{enumerate}
\item We unfreeze the entire ViT and report the experimental results the last two rows in Table \ref{tab:special-comp}. This setup leads to severe overfitting.
\item We implement a ResNet version of InfLoRA. As seen in the first two row with asterisk in Table \ref{tab:special-comp}, it fails to match our method and other regular ones in Table \ref{tab:method-benchmark}, even when initializing from a pretrained ResNet model.
\end{enumerate}

The former confirms that ViT's pretrained knowledge is crucial to its performance, while the latter suggests that InfLoRA is ineffective when this advantage is removed. These results indicate that freezing the backbone while tuning LoRA modules is a scenario-specific optimization rather than a general incremental learning solution, reinforcing the advantages of AdaLL over InfLoRA.

\paragraph{Hyperparameter Sensitivity}

We analyze AdaLL’s stability under varying bottleneck widths, selecting EWC \citep{kirkpatrick2017overcoming} and LwF \citep{li2017LwF} as baselines representing weight- and prediction-based regularization methods, respectively.

As shown in Figure \ref{fig:param_comp}, weight-regularized models cluster at the top, except for the setup with width 32. Prediction-based methods, though forming two distinct groups, exhibit strong consistency, with two configurations ranking highest with minimal variation.

AdaLL’s moderate sensitivity to hyperparameters allows tuning for optimal performance while maintaining stability, as the optimal setting is not unique.

\section{Conclusion}
We propose AdaLL, a framework that leverages adapters to address the stability-plasticity dilemma. Unlike other task-specific methods, AdaLL incorporates co-training with regularization constraints, offering a solution that is both simple and effective. Its simplicity lies in reducing the problem to a backbone-adapter uni-directional structure, where the backbone and adapters independently contribute to stability and plasticity, which is further enforced by the backbone regulariztion algorithms. Its effectiveness is demonstrated through extensive experimental results, including analyses on co-training, regularization, and comparisons with both traditional and modern algorithms.  

Our results show that AdaLL not only outperforms all counterparts but also revitalizes traditional algorithms, elevating them back to competitive benchmarks. Additionally, it exhibits exceptional compatibility by integrating with non-adapter algorithms and further enhancing their performance. 

\clearpage
{
    \small
    \bibliographystyle{ieeenat_fullname}
    \bibliography{main}
}


\clearpage

\begin{figure*}[t!]
    \centering
    \includegraphics[width=0.9\textwidth]{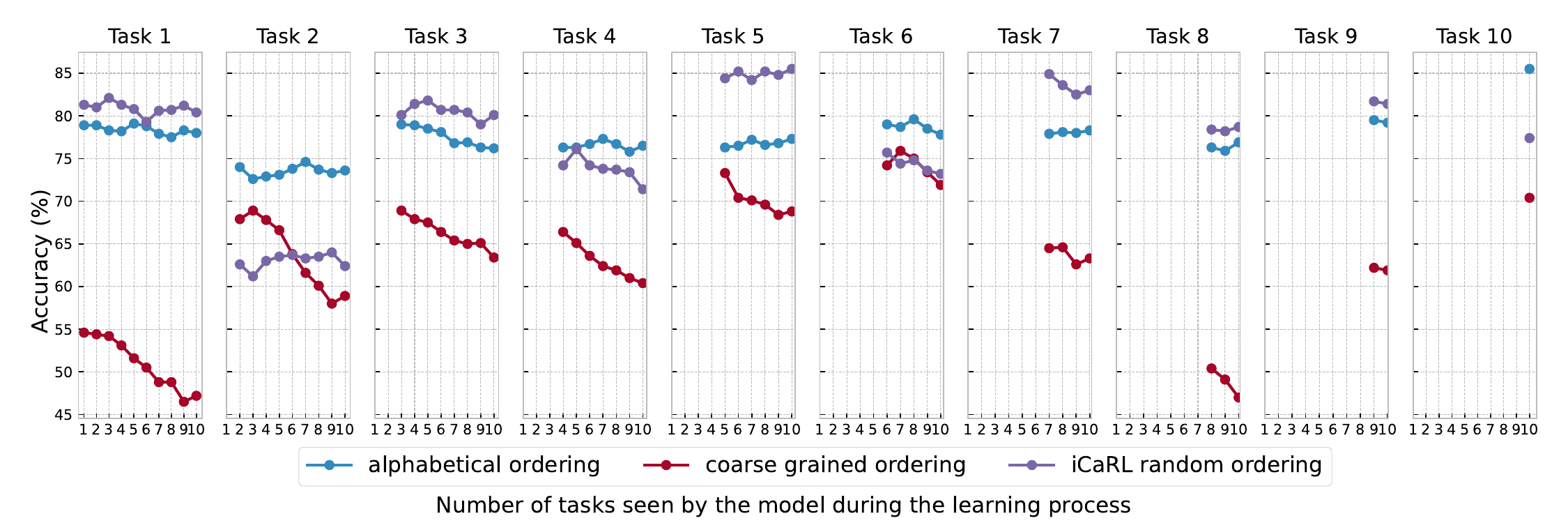}
    \caption{The model’s accuracy using LwF for incremental learning is evaluated on three different CIFAR-100 task orderings: the standard alphabetical category ordering, the coarse grained ordering, and the iCaRL \citep{rebuffi2017icarl} ordering. The coarse grained ordering has more inter-task diversity.} 
    \label{fig:fig_lwf_task_order}
\end{figure*}

\section{The Importance of Task Ordering}\label{apdx:task-ordering}

Figure \ref{fig:fig_lwf_task_order} examines the impact of task orderings on the classification performance of the LwF algorithm applied to CIFAR-100. In this scenario, the model is incrementally trained with 10 classes per task over a total of 10 tasks. Three different task orderings are evaluated: an alphabetical class ordering, a random ordering with a fixed seed commonly used by iCaRL and other methods \citep{rebuffi2017icarl, zhang2020class}, and a coarse grained ordering that groups similar classes within each task based on CIFAR-100’s 20 coarse categories. 

The coarse grained ordering has greater inter-task diversity, providing a way to assess how increased inter-task differences affect the incremental learning algorithm. Notably, when classes are learned in a coarse-grained ordering, there is a significant increase in both forgetting and accuracy loss for each task compared to other orderings. This suggests that algorithms behave differently when different orderings are presented. Since previous works generally overlook the importance of this, we include the relevant evaluations of our framework in Section \ref{sec:comprehensive-comparison}.

\section{Implementation details}\label{appendix:sec:implementation}

\begin{table*}[h!]
\centering
\begin{tabular}{|l|c|c|c|c|c|}
\hline \text { Datasets } & \char"0023 \text { Train } & \char"0023 \text { Validation } & \char"0023 \text { Test } & \text { Input size } & \text {Batch size} \\
\hline \hline \text { CIFAR-100 } & 45,000 & 5,000 & 10,000 & $32 \times 32 \times 3$ & 128 \\
\text { ImageNet-Subset } & 117,000 & 13,000 & 5,000 & $224 \times 224 \times 3$ & 4 \\
\hline
\end{tabular}
\caption{Summary of datasets used. Both datasets contain 100 classes each.}
    \label{tab:datasets}
\end{table*}

\begin{table*}[h!]
\centering
\begin{tabular}{|c|c|c|c|c|c|c|}
\hline \text{ hyperparameter} & EWC & MAS & PathInt & LwF & LwF-reg & LwM\\
\hline $\lambda$ & 10,000 & 400 & 10 & 10 &5+1 & 2 \\
\hline
\end{tabular}
\caption{Summary of hyperparameter to control the stability and plasticity used across regularization-based methods. For the LwF with adapter and regularizations (LwF-reg), the two $\lambda$'s ($\lambda_{distill}, \lambda_\varphi$) are set to 5 and 1 respectively.}
    \label{tab:hyperparams}
\end{table*}

We follow the framework FACIL \citep{masana2022class_survey}, and code is implemented using Pytorch. We apply SGD with momentum set to 0.9 and weight decay set to 0.0002.

\subsection{Datasets}\label{appendix:sec:dataset}

Table \ref{tab:datasets} presents a summary of the datasets used in the experiments. We apply data augmentation to both datasets, including padding, cropping, input normalization, and random horizontal flipping. Specifically, for CIFAR-100, we apply a padding of 4 pixels to each side of the image, followed by random cropping to $32 \times 32$ for training purposes and center cropping for testing. We apply the same but resize the image to $224 \times 224$ for training ViT. For the ImageNet-Subset, we resize the images to $256 \times 256$, then perform random cropping to $224 \times 224$ for training and center cropping for testing.

\subsection{Learning rate search and hyperparameter tuning} \label{appendix:sec:hyper}

We apply the Continual hyperparameter Framework \citep{de2021continual}, a common framework to select learning rates (LRs) and tune hyperparameters for methods of incremental learning.  The framework incorporates two phase: the Maximal Plasticity Search phase to search LR with fine-tuning on the new task, and the Stability Decay phase to search for the optimal hyperparameters.

For the Maximal Plasticity Search phase, we fine-tune the model on the new task and select the optimal LR to achieve high plasticity. Specifically, we train the model from scratch for the first task and apply LR search on \{5e-1, 1e-1, 5e-2\}. Starting from the second task, the LR search space is limited to {1e-1, 5e-2, 1e-2, 5e-3, 1e-3\}. LR decay is applied, with a decay factor of 3 and patience of 10 epochs. The stopping criteria is either the LR below 1e-4 or 100 epochs have passed.  

For the Stability Decay phase, we fix the LR and select the hyperparameter starting from a high value, with a gradual decay to achieve the optimal stability-plasticity trade-off. We start from a high value of hyperparameter because this is close to freezing the network such that old knowledge is preserved, and through gradual decay the model becomes less intransigence and slowly converges towards higher forgetting, which would ultimately corresponds to fine-tuning of the previous phase. Specifically, we reduce the hyperparameter value in half if the method accuracy is below the 95\% of the fine-tuning accuracy in the previous LR search phase. 

Next, we present the starting values for hyperparameters across methods. Most of the methods discussed in our paper are regularization-based approach, which include a hyperparameter $\lambda$ for the regularizer that controls the trade-off between stability and plasticity. The starting values for $\lambda$ are summarized in Table \ref{tab:hyperparams}. For other method-specific parameters, we generally follow the corresponding original work. For instance, we fix the temperature parameter as 2 for LwF. For PathInt, the dampling parameter is set to 0.1. For LwM, the attention distillation parameter is set to 1 based on an empirical evaluation in \citep{masana2022class_survey}. 

Due to computational cost constraints, we have adjusted the stopping criteria for the ImageNet Subset to 80 epochs, rather than the previously mentioned 100. Additionally, we do not conduct LR searches or tune hyperparameters for this dataset, and fix the LR at 0.01.

\subsection{Special Implementation for InfLoRA, iTAML, and DualPrompt} \label{appendix:sec:implementation-detail}
iTAML, InfLoRA, and DualPrompt utilize method-specific setups and hyperparameters without additional hyperparameter search. We implement them using their default configurations as reported in the original works to ensure alignment with their optimal settings. The details are as follows:

\begin{itemize}
    \item \textbf{iTAML:} We use a ResNet-18 trained for 70 epochs with a memory size of 3000, $\mu=1$, $\beta=2$, and $r=5$. The learning rate is set to $0.01$.
    \item \textbf{DualPrompt:} We run a ViT-Base-Patch16-224 with both G-prompt and E-prompt enabled. The model is optimized using Adam with a learning rate of $0.03$, and each task is trained for 2 epochs.
    \item \textbf{InfLoRA:} We evaluate different configurations, including the default one in Table \ref{tab:method-benchmark} and the rest two in Table \ref{tab:special-comp}:
    \begin{itemize}
        \item Default: Learning rate of $3\text{e-}2$, LoRA rank 10, and all required backbone parameters frozen.
        \item ViT$^\dagger$: Identical to default but with all parameters unfrozen except LoRA modules, ensuring task-specific activations as in the original work.
        \item ResNet*: Uses a ResNet backbone with required weights frozen. The ResNet LoRA is built by connecting the input and the output of each convolution layer. The representations are flattened and resized when needed.
    \end{itemize}
\end{itemize}

For a fair comparison, we use LwF-A as the best-performing AdaLL variant and apply it to the different InfLoRA setups for comparison. In these experiments, we strictly match the LwF-A configurations with InfLoRA to eliminate confounding factors.

\subsection{The Impact of Gradient Clipping}

During our implementation, we found gradient clipping to be a critical factor for incremental learning algorithms, with the optimal clipping value varying across methods. Figure \ref{fig:clipping} illustrates the significant impact of different gradient clipping values on model performance, particularly from the very first task.

Based on this observation, we apply method-specific gradient clipping for adapters in our universality experiments. Specifically, for weight-regularized methods such as EWC and MAS, we use a clipping value of 10, whereas for prediction-regularized methods, we use a clipping value of 1. This also explains why, in Figure \ref{fig:fig_data_scale}, the performance curves of EWC and LwF adapters do not align at the beginning of the experiments.

\begin{figure}[h]
    \centering
    \includegraphics[width=0.35\textwidth]{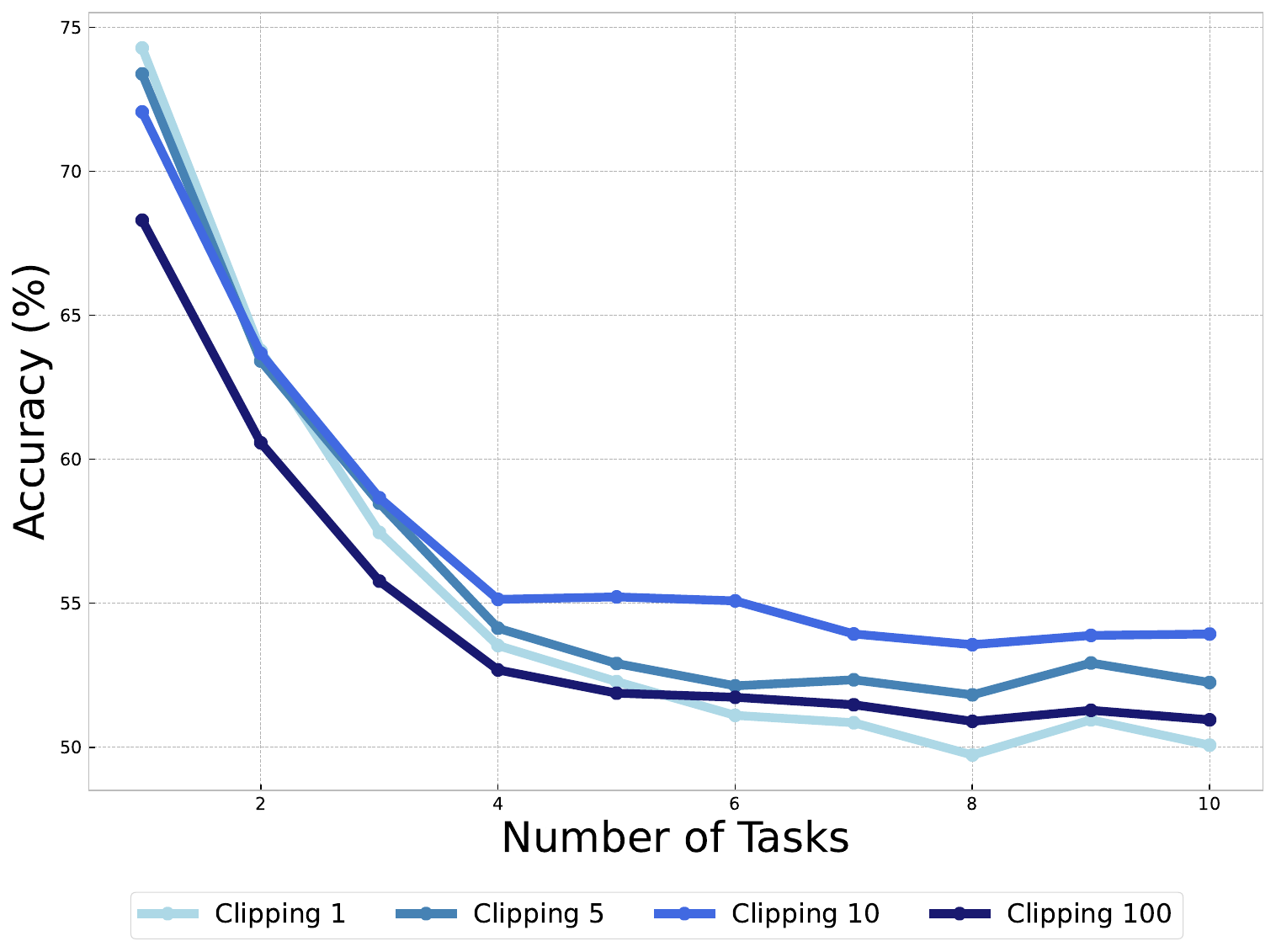}
    \caption{The average accuracy for EWC-A with different gradient clipping values on CIFAR-100 in Task-IL.}
    \label{fig:clipping}
\end{figure}

\section{Class-IL results} \label{appendix:sec:tag}

Figure \ref{fig:ordering_comp_tag} illustrates the average accuracy of all methods in the setup described in Section \ref{sec:comprehensive-comparison}, specifically for class incremental learning (Class-IL). Figure \ref{fig:fig_data_scale_tag} demonstrates the results across various task scales, while Figure \ref{fig:ordering_comp_tag} presents the results based on different orderings. Figure \ref{fig:imgnet-tag} provides the results for ImageNet.

\begin{figure*}[h]
    \centering
    \includegraphics[width=\textwidth]{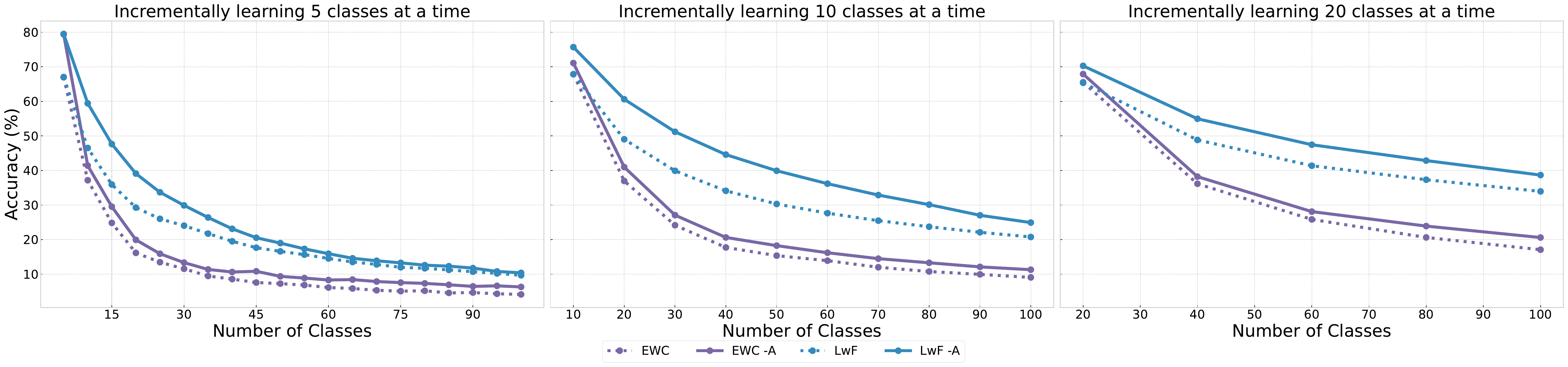}
    \caption{The average accuracy for EWC and LwF with learning 5, 10, and 20 classes at a time on CIFAR-100 (alphabetical ordering) in Class-IL. }
    \label{fig:fig_data_scale_tag}
\end{figure*}

\begin{figure*}[h]
    \centering
    \includegraphics[width=\textwidth]{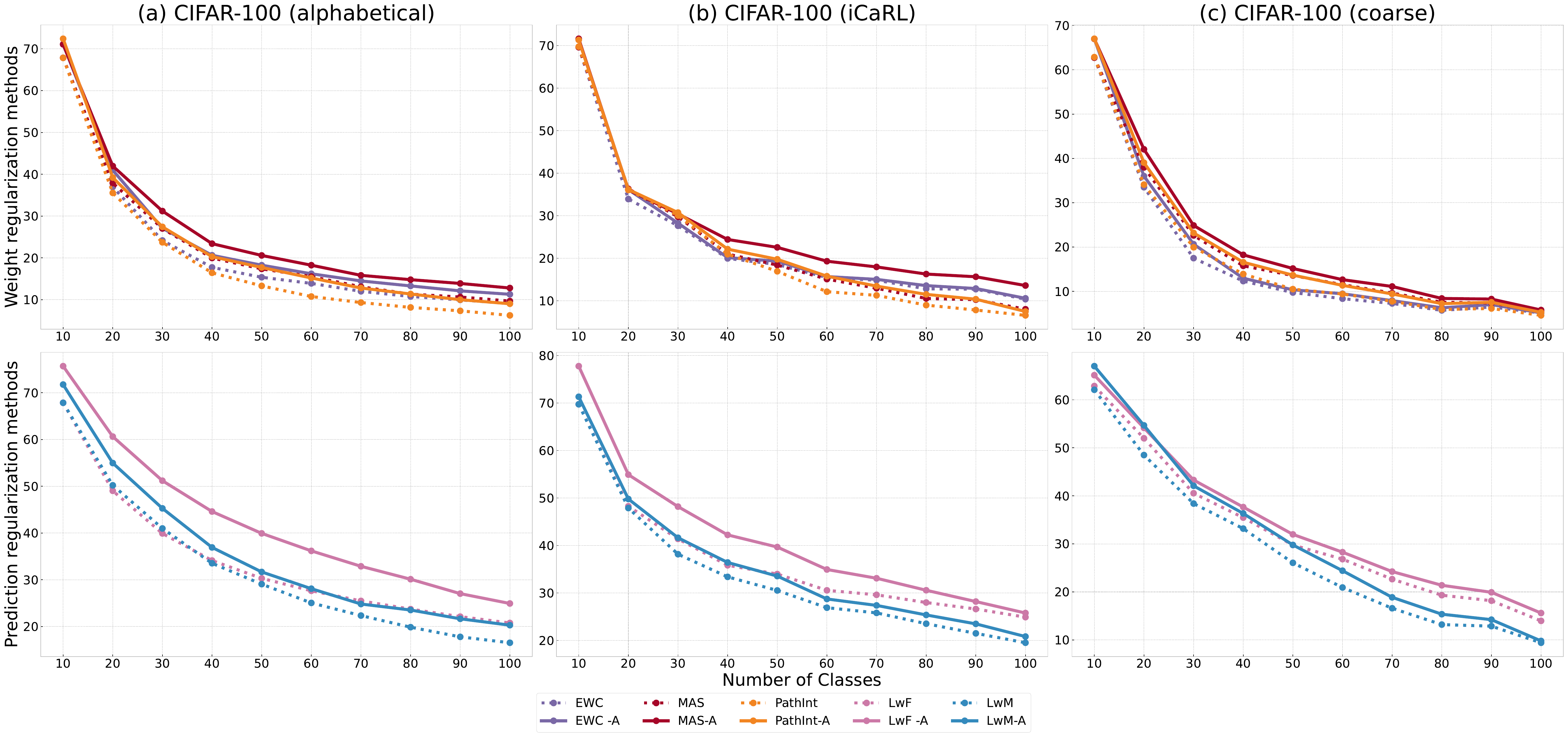}
    \label{fig:ordering_cifar_tag}
    \caption{The average accuracy for regularization-based methods with or without adapters on CIFAR-100 (all orderings) in Class-IL. The solid line represents the results with adapter, while the dashed line represents the results without adapter.}
    \label{fig:ordering_comp_tag}
\end{figure*}
\begin{figure*}[t]
    \centering
    \includegraphics[width=0.8\textwidth]{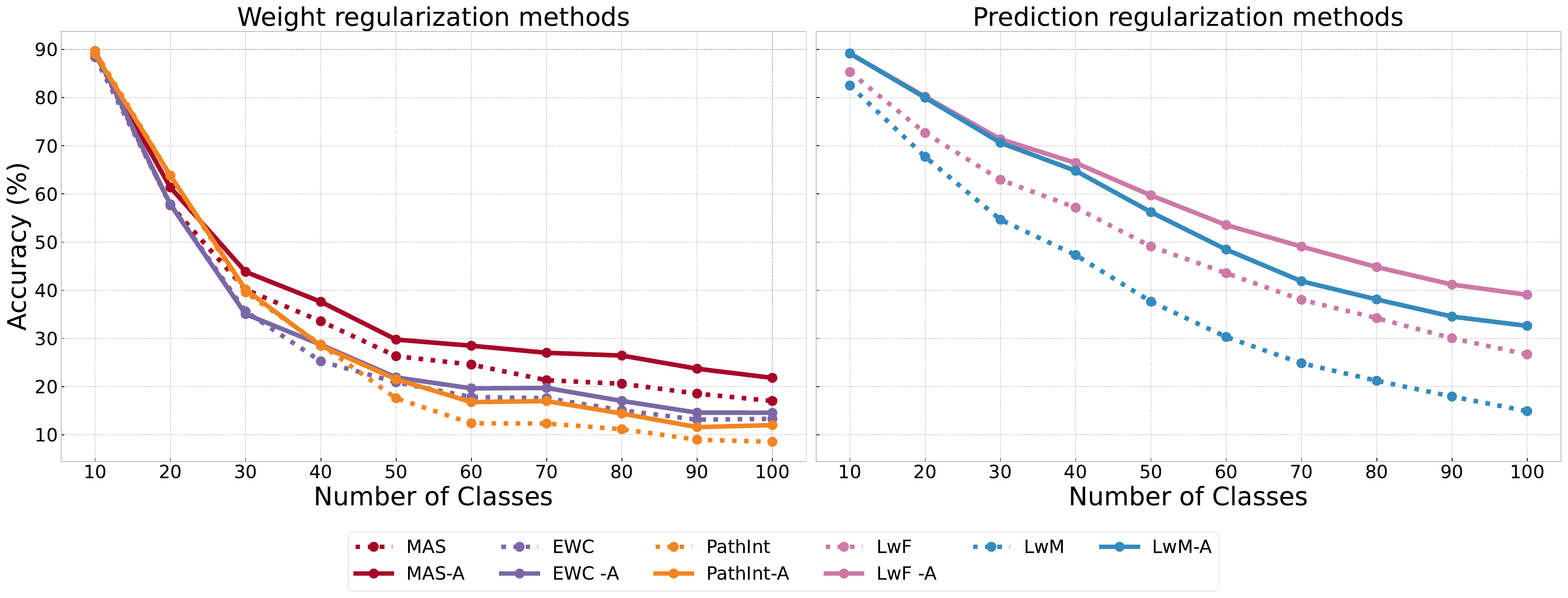}
    \caption{The average accuracy for regularization methods on ImageNet-subset. The x-axis represent the number of tasks and the y axis represents the TOP-1 accuracy (\%). The solid line represents the results with adapter, while the dashed line represents the results without adapter.}
    \label{fig:imgnet-tag}
\end{figure*}

\end{document}

%% file: preamble.tex
\usepackage[utf8]{inputenc}

\usepackage{minitoc}

\def\$#1\${\begin{align*}#1\end{align*}}